\documentclass{article}

\usepackage{microtype}
\usepackage{graphicx}
\usepackage{csquotes}
\usepackage{multirow}     % \multirow
\usepackage{pifont}
\usepackage{tikz}
\usepackage{url}
\usepackage{pgfplots}
\usepgfplotslibrary{polar}
\pgfplotsset{compat=1.17}
\usepackage[most]{tcolorbox}
\usepackage{makecell}     
\usepackage{listings}
\tcbuselibrary{listings} 
\usepackage[table,dvipsnames]{xcolor} 
\definecolor{myblue}{RGB}{31, 119, 180}
\definecolor{myorange}{RGB}{255, 127, 14}
\definecolor{mygreen}{RGB}{44, 160, 44}
\definecolor{row-blue}{HTML}{F5FFFA}
\definecolor{row-green}{HTML}{F1F6EC}
\definecolor{row-yellow}{HTML}{FFFFF0}
\definecolor{row-pink}{HTML}{FFF5F5}
\definecolor{superlightgrey}{gray}{0.7}
\definecolor{kellygreen}{rgb}{0.3, 0.73, 0.09}
\definecolor{alizarin}{rgb}{0.82, 0.1, 0.26}
\definecolor{primaryColor}{HTML}{2E86AB} 
\definecolor{accentColor}{HTML}{D9534F}   
\definecolor{bgColor}{HTML}{F7F9FB}       
\definecolor{codeBg}{HTML}{EDF2F7}      
\definecolor{textColor}{HTML}{2D3748}     

\definecolor{cmarkgreen}{RGB}{50, 168, 82} 
\definecolor{xmarkred}{RGB}{219, 50, 54}    
\definecolor{rowgray}{gray}{0.95}           

\newcommand{\cmark}{\textcolor{cmarkgreen}{\ding{51}}}%
\newcommand{\xmark}{\textcolor{xmarkred}{\ding{55}}}%
\newcommand{\gray}[1]{\textcolor{gray}{#1}}
\newcommand{\best}[1]{\textbf{#1}}
\newcommand{\second}[1]{\underline{#1}}
\definecolor{improvementcolor}{HTML}{009999} 
\newcommand{\gain}[1]{\textcolor{improvementcolor}{\scriptsize{~$\uparrow$#1}}}
\usepackage{arydshln}    

\usepackage{graphicx}    

\usepackage{bm}          
\usepackage{amsmath,amssymb} 

\newcolumntype{I}{!{\vrule width 1.2pt}}

\usepackage{subcaption}
\usepackage{booktabs} 
\usepackage{colortbl}
\usepackage{array}
\usepackage{fontawesome5}
\usepackage{caption}
\usepackage{tabularx}

\usepackage{hyperref}

\usepackage[preprint]{icml2026}

\usepackage{amsmath}
\usepackage{amssymb}
\usepackage{mathtools}
\usepackage{amsthm}

\usepackage[capitalize,noabbrev]{cleveref}

\theoremstyle{plain}
\newtheorem{theorem}{Theorem}[section]
\newtheorem{proposition}[theorem]{Proposition}

\theoremstyle{definition}
\newtheorem{definition}[theorem]{Definition}

\theoremstyle{remark}

\usepackage[textsize=tiny]{todonotes}

\icmltitlerunning{AD-MIR: Bridging the Gap from Perception to Persuasion in Advertising Video Understanding via Structured Reasoning}

\begin{document}

\twocolumn[
  \icmltitle{AD-MIR: Bridging the Gap from Perception to Persuasion in Advertising Video Understanding via Structured Reasoning}

  \icmlsetsymbol{equal}{*}
  \icmlsetsymbol{leader}{†}
  \icmlsetsymbol{corr}{$\dagger$} 

  \begin{icmlauthorlist}
    \icmlauthor{Binxiao Xu}{pku,xjtu,equal} 
    \icmlauthor{Junyu Feng}{xjtu,equal}
    \icmlauthor{Xiaopeng Lin}{xjtu}
    \icmlauthor{Haodong Li}{south}
    \icmlauthor{Zhiyuan Feng}{thu}
    \icmlauthor{Bohan Zeng}{pku} \\
    \icmlauthor{Shaolin Lu}{pku,leader} 
    \icmlauthor{Ming Lu}{intel}
    \icmlauthor{Qi She}{Bytedance}
    \icmlauthor{Wentao Zhang}{pku,corr}
    %\icmlauthor{}{sch}
    %\icmlauthor{}{sch}
    %\icmlauthor{}{sch}
  \end{icmlauthorlist}

  \icmlaffiliation{pku}{Peking University, Beijing, China}
  \icmlaffiliation{thu}{Tsinghua University, Beijing, China}
  \icmlaffiliation{xjtu}{Xi'an Jiaotong University, Xi'an, China}
  \icmlaffiliation{south}{South China University of Technology, Guangzhou, China}
  \icmlaffiliation{intel}{Intel, Beijing, China}
  \icmlaffiliation{Bytedance}{ByteDance, Beijing, China}
  
  \icmlcorrespondingauthor{Wentao Zhang}{wentao.zhang@pku.edu.cn}

  % You may provide any keywords that you find helpful for describing your
  % paper; these are used to populate the "keywords" metadata in the PDF but
  % will not be shown in the document
  \icmlkeywords{Machine Learning, ICML}

  \vskip 0.3in
]

% this must go after the closing bracket ] following \twocolumn[ ...

% This command actually creates the footnote in the first column listing the
% affiliations and the copyright notice. The command takes one argument, which
% is text to display at the start of the footnote. The \icmlEqualContribution
% command is standard text for equal contribution. Remove it (just {}) if you
% do not need this facility.

% Use ONE of the following lines. DO NOT remove the command.
% If you have no special notice, KEEP empty braces:
% \printAffiliationsAndNotice{}  % no special notice (required even if empty)
% Or, if applicable, use the standard equal contribution text:
% \printAffiliationsAndNotice{\icmlEqualContribution}
\printAffiliationsAndNotice{
  \icmlEqualContribution
 \textsuperscript{†}~Project~leader
  \textsuperscript{\textdagger}~Corresponding~author
}

\begin{abstract}
Multimodal understanding of advertising videos is essential for interpreting the intricate relationship between visual storytelling and abstract persuasion strategies. However, despite excelling at general search, existing agents often struggle to bridge the cognitive gap between pixel-level perception and high-level marketing logic. To address this challenge, we introduce \textbf{AD-MIR}, a framework designed to decode advertising intent via a two-stage architecture. First, in the \textbf{Structure-Aware Memory Construction} phase, the system converts raw video into a structured database by integrating semantic retrieval with exact keyword matching. This approach prioritizes fine-grained brand details (e.g., logos, on-screen text) while dynamically filtering out irrelevant background noise to isolate key protagonists. Second, the \textbf{Structured Reasoning Agent} mimics a marketing expert through an iterative inquiry loop, decomposing the narrative to deduce implicit persuasion tactics. Crucially, it employs an evidence-based self-correction mechanism that rigorously validates these insights against specific video frames, automatically backtracking when visual support is lacking. Evaluation on the AdsQA benchmark demonstrates that AD-MIR achieves state-of-the-art performance, surpassing the strongest general-purpose agent, DVD, by 1.8\% in strict and 9.5\% in relaxed accuracy. These results underscore that effective advertising understanding demands explicitly grounding abstract marketing strategies in pixel-level evidence. The code is available at \url{https://github.com/Little-Fridge/AD-MIR}.
\end{abstract}

\section{Introduction}

The research paradigm for video understanding is shifting from entity-centric perception to intent-oriented cognition, as \citet{Lin2024,Zhang2025,Fu2025} progressively move from object-centric recognition to higher-level reasoning over long-form video content. Despite the success of Large Multimodal Models (LMMs) in generalizing across objective physical descriptions, a substantial cognitive disparity remains when processing subjective and strategic content, even as \citet{Liu2023,Zhang2024_LongVA} demonstrate strong generalization under instruction-tuned and long-context settings. Advertising videos serve as an adversarial benchmark for this limitation. Unlike objective recordings, ads are engineered semiotic systems \citep{Hussain2017}. In such non-linear narratives, cinematic techniques such as lighting manipulation and rhythmic editing function not merely as visual signals, but as vehicles for abstract persuasion logic. Therefore, bridging this semantic divide between low-level visual facts (e.g., a scene fading to black) and high-level abstract intents (e.g., creating a sense of suppression to trigger pain points) is essential for effective advertising understanding.

\begin{table}[htbp]
    \centering
    \caption{\textbf{Feature comparison with existing paradigms.} Unlike general agents (e.g., DVD) limited to generic retrieval, or RL-optimized baselines (ReAd-R) relying on implicit outcome fitting, AD-MIR explicitly integrates expert modules to decode persuasion strategies ("why") and employs rigorous visual verification to minimize hallucinations.}
\resizebox{\columnwidth}{!}{
    \begin{tabular}{l c c c c c}
        \toprule
        & \multicolumn{2}{c}{\textbf{Perception}} & \multicolumn{2}{c}{\textbf{Strategy Cognition}} & \textbf{Reliability} \\
        \cmidrule(lr){2-3} \cmidrule(lr){4-5} \cmidrule(lr){6-6}
        \textbf{Method} & 
        \begin{tabular}{@{}c@{}}Visual \\ Grounding\end{tabular} & 
        \begin{tabular}{@{}c@{}}Fine-grained \\ Detail\end{tabular} & 
        \begin{tabular}{@{}c@{}}Causal \\ Reasoning\end{tabular} & 
        \begin{tabular}{@{}c@{}}Persuasion \\ Decoding\end{tabular} &  
        \begin{tabular}{@{}c@{}}Visual \\ Verification\end{tabular} \\
        \midrule
        
        End-to-End LMMs & \xmark & \cmark & \xmark & \xmark & \xmark \\
        
        General Video Agents (DVD) & \cmark & \cmark & \xmark & \xmark & \xmark \\

        ReAd-R (AdsQA Baseline) & \cmark & \cmark & \cmark & \xmark & \xmark \\ 
        
        \rowcolor{rowgray}
        \textbf{AD-MIR (Ours)} & \cmark & \cmark & \cmark & \cmark & \cmark \\
        \bottomrule
    \end{tabular}
    }
    \label{tab:comparison}
\end{table}

\begin{figure*}[t]
  \begin{center}
    \centerline{\includegraphics[width=\textwidth]{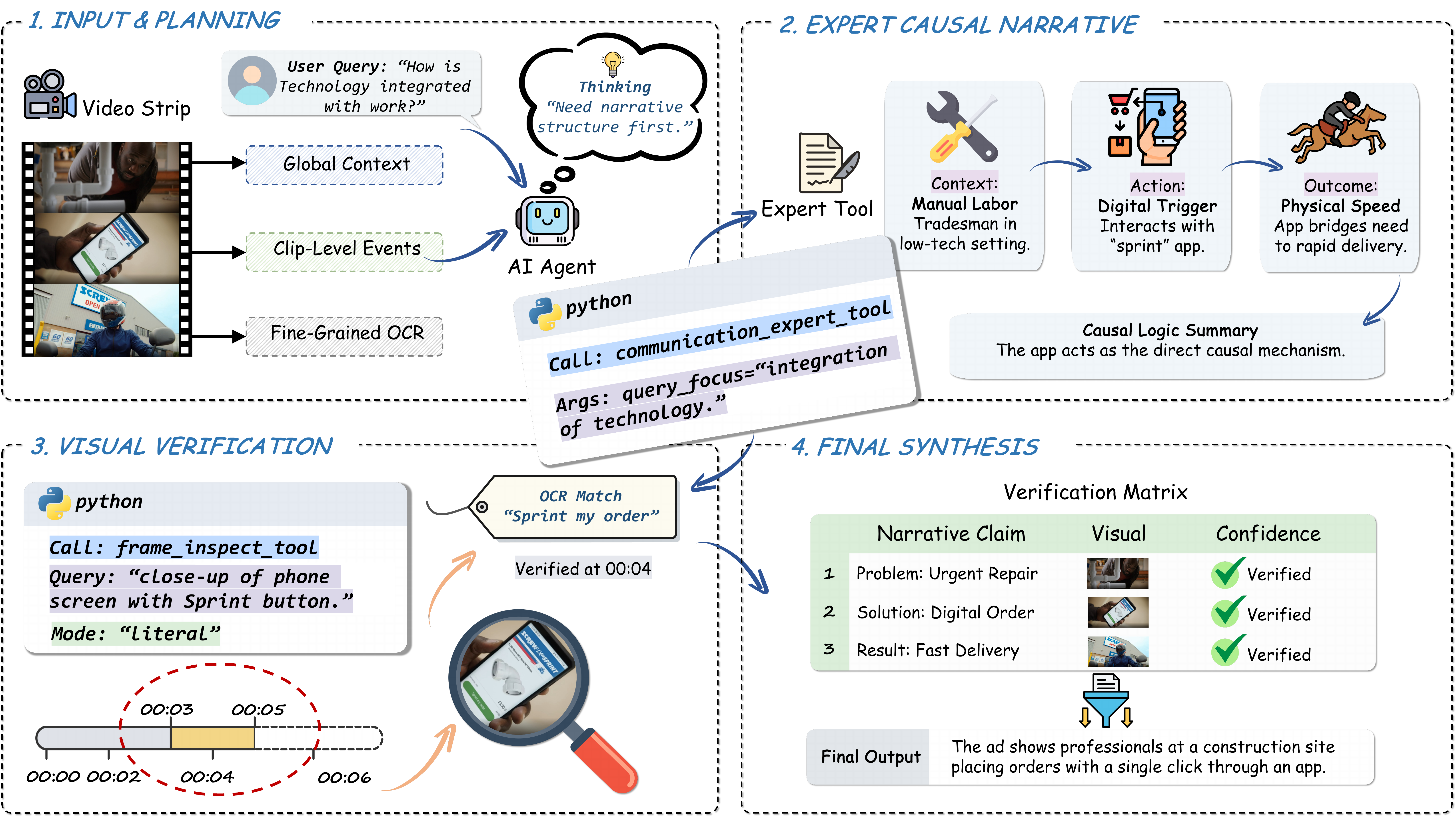}}
\caption{\textbf{Illustrative walkthrough of AD-MIR’s reasoning pipeline on an AdsQA example.} Unlike general agents that focus on retrieval, AD-MIR bridges the cognitive gap by first constructing a high-level causal narrative via a communication expert (Phase 1-2), then performing targeted frame inspection to verify precise visual details (Phase 3), and finally synthesizing a visually grounded explanation that links the expert narrative to pixel-level evidence (Phase 4).}
    \label{fig:demo}
  \end{center}
\end{figure*}

Current video understanding models face significant hurdles in the domain of advertising \citep{Hussain2017}. While end-to-end LMMs excel at describing general visual content \citep{Liu2023,Lin2024}, they struggle with two main limitations: limited memory and logic hallucinations \citep{Zhang2024_LongVA,Guan2024,Fu2025}. First, constrained by context windows, these models often lose track of details across long narratives. Second, and more critically, lacking a mechanism to decode persuasive intent, they tend to hallucinate when explaining strategic motivations, often fabricating reasons that defy basic advertising logic \citep{Hussain2017,Guan2024}.

On the other hand, tool-augmented agents, represented by the Deep Video Discovery (DVD) Agent \citep{Zhang2025}, have successfully established a baseline for \textit{visual grounding} by precisely locking onto the timestamps of events. However, a critical distinction exists between retrieving \textit{"what happened"} and reasoning \textit{"why it happened."} While DVD demonstrates acuity in locating visual content, it lacks the cognitive capability to decode the underlying marketing psychology(e.g., identifying rhetorical devices, emotional appeals, and symbolic metaphors). Even the recent ReAd-R framework \citep{Long2025AdsQA}, which employs reinforcement learning to enhance retrieval-QA alignment, remains limited by implicit data fitting. Without explicit structural constraints for persuasion analysis, such models struggle to distinguish between superficial visual correlations and deep strategic causation (e.g., recognizing a phone not merely as an object, but as a solution to a labor problem). Consequently, when nuanced strategic inquiries arise, standard video agents and RL-optimized baselines can only draw shallow associations, resulting in a logical fracture between visual perception and intent understanding. \cref{tab:comparison} summarizes the capability gaps between these existing paradigms and our proposed approach.

In response to these challenges, we propose \textbf{AD-MIR} (\underline{AD}vertising \underline{M}ultimodal \underline{I}ntent \underline{R}easoning), a multimodal agent framework tailored for advertising videos. Rather than relying on single-shot retrieval or black-box inference, AD-MIR is designed to close the cognitive gap via a two-stage architecture that extends the standard reasoning-and-acting paradigm \citep{Yao2023}. As illustrated in \cref{fig:demo}, the framework operates by mirroring how a human analyst works: first, constructing a structured memory of the narrative, and second, employing an iterative inquiry loop to align high-level hypotheses with precise visual anchors.

The first phase is \textit{Structure-Aware Memory Construction}, where the system decouples raw video streams into visual, textual, and audio modalities to construct a hierarchical index. This allows the model to anchor cues often missed by general retrieval. The second phase involves the \textit{Structured Reasoning Agent}, which employs an explicit iterative inquiry-refinement mechanism. Instead of a generic loop, this mechanism intelligently selects the most suitable tools based on the query's nature. For factual queries, it routes to perception tools; for strategic queries, it invokes specialized communication expert modules grounded in marketing psychology (e.g., analyzing persuasion, rhetoric, and emotional arcs). This hierarchical synergy enables AD-MIR to capture implicit logical associations, such as linking \enquote{a plumber under a sink} to \enquote{a speeding motorbike} by reconstructing discrete visual symbols into a coherent marketing narrative.

Our main contributions are summarized as follows:
\begin{enumerate}
\item \textbf{A Multimodal Reasoning Agent Framework Tailored for Advertising.} We present AD-MIR, a novel agent designed to bridge the gap between visual perception and intent understanding. By employing an iterative inquiry-refinement mechanism that aligns expert-derived marketing insights with rigorous visual verification, it effectively decodes complex persuasion tactics that general agents miss.

\item \textbf{Hierarchical Perception-Inference Toolchain.} We introduce a query-driven architecture that selectively routes inquiries between perception tools for factual grounding and specialized expert modules grounded in marketing psychology. This specialized design enables the precise interpretation of symbolic imagery and emotional cues, surpassing the literal description capabilities of standard multimodal systems.

\item \textbf{Visual Verification for Grounded Reasoning.} We introduce a mechanism that validates high-level reasoning against pixel-level evidence, ensuring interpretations are derived strictly from video frames rather than linguistic priors. This cross-modal alignment bridges abstract marketing concepts with concrete visual signals, significantly enhancing the reliability and interpretability of persuasion analysis.

\end{enumerate}

\begin{figure*}[t]
  \begin{center}
\centerline{\includegraphics[width=\textwidth]{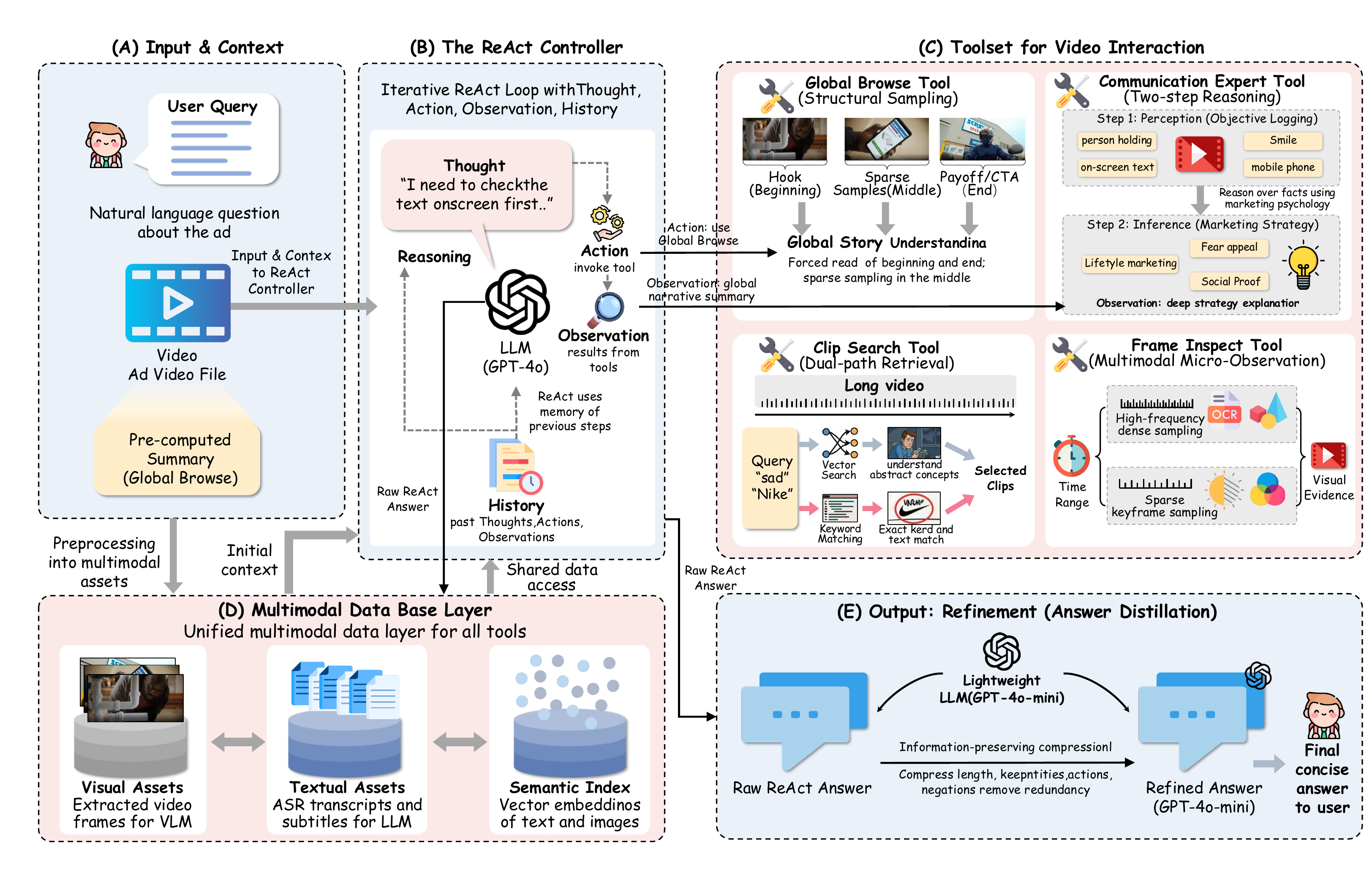}}
\caption{\textbf{The overall architecture of AD-MIR.} The framework comprises five synergistic components: (A) Input \& Context construction for multi-modal preprocessing; (B) a ReAct Controller for iterative reasoning; (C) a Hierarchical Toolset featuring global browsing, communication experts, and fine-grained inspection; (D) a Unified Multimodal Database serving as shared memory; and (E) an Answer Refinement stage to ensure concise, evidence-based output.}
    \label{fig:arch}
  \end{center}
\end{figure*}

\section{Method}

In this section, we present the AD-MIR framework. As illustrated in \cref{fig:arch}, AD-MIR utilizes an agentic reasoning architecture that couples a ReAct-style controller with a shared multimodal database $\mathcal{M}$ and a suite of domain-specific interaction tools. This design effectively bridges the semantic gap between pixel-level perception and high-level marketing intent. The workflow proceeds in two synergistic stages: (1) \textbf{Structure-Aware Memory Construction}, where raw video streams are decoupled into a discrete, indexable multimodal database to isolate key narrative elements; and (2) \textbf{Iterative Intent Reasoning}, where a domain-adaptive agent orchestrates perception primitives and intent experts via a \enquote{Think-Act-Observe} loop, aligning visual evidence with logically grounded marketing inference.

\subsection{Problem Definition}

We formalize the advertising video understanding task as a Partially Observable Markov Decision Process (POMDP), defined by the tuple $(\mathcal{S}, \mathcal{A}, \mathcal{T}, \mathcal{R}, \Omega, \mathcal{O})$ \cite{Wu2019AdaFrame}. Due to the non-linear and metaphorical nature of advertising narratives, the agent cannot perceive the full video $\mathcal{V}$ in a single step, rendering the state space $\mathcal{S}$ partially observable.

\begin{definition}
  \label{def:state_space}
  The state $s_t \in \mathcal{S}$ at time $t$ is defined as the union of the multimodal database $\mathcal{M}$, the context-anchored subject registry $\mathcal{S}_{reg}$, and the interaction history $\mathcal{H}_{t-1}$. 
\end{definition}

Given a natural language query $Q$, the agent receives an environmental observation $o_t \in \Omega$. It then generates a Chain-of-Thought (CoT) reasoning step $z_t$ by leveraging the history $\mathcal{H}_{t-1} = \{o_0, a_0, \dots, a_{t-1}, z_{t-1}\}$.

\begin{proposition}
  \label{prop:policy}
  The agent selects the next action $a_t \in \mathcal{A}$ according to a domain-adaptive policy $\pi(a_t | \mathcal{H}_{t-1}, o_t; \Theta)$, where $\Theta$ represents the frozen parameters of a Large Multimodal Model. The policy is implemented via prompt-based in-context learning rather than gradient-based updates.
\end{proposition}

The system aims to generate an answer $A^*$ that maximizes semantic consistency and evidence-based grounding:
\begin{equation}
A^* = \arg\max_{A}P(A | \mathcal{V}, Q, \mathcal{S}_{reg}; \Theta)
\end{equation}

\subsection{Structure-Aware Memory Construction}
To effectively mitigate the computational VRAM bottlenecks inherent in long-video reasoning and simultaneously ensure the granular precision of evidence backtracking, we employ a dedicated offline pipeline to decouple the continuous raw video stream into a structured multimodal database. This systematic preprocessing workflow is primarily underpinned by two pivotal mechanisms:

\subsubsection{Hybrid Semantic-Lexical Indexing} 
We discretize the raw video stream into a set of temporal clips $\mathcal{C} = \{c_i\}_{i=1}^N$. To construct a comprehensive semantic index, for each clip $c_i$, we utilize a VLM to generate fine-grained visual narrative descriptions $d_{\text{vis}}^{(i)}$ and integrate the Whisper model to extract timestamped audio transcripts $d_{\text{aud}}^{(i)}$. Addressing the precision loss inherent in single dense vector retrieval when handling specific entities (e.g., brand names, prices), we propose a hybrid semantic-lexical scoring strategy. Given a query $Q$, the relevance score $S(Q, c_i)$ for clip $c_i$ is formalized as a weighted linear combination of dense semantic similarity and exact keyword matching:
\begin{equation}
\resizebox{0.9\linewidth}{!}{$
S(Q, c_i) = \underbrace{\cos(\phi(Q), \phi(d_{\text{vis}}^{(i)} \oplus d_{\text{aud}}^{(i)}))}_{\text{Semantic}} + \beta \cdot \underbrace{\mathcal{K}(Q, d_{\text{vis}}^{(i)} \oplus d_{\text{aud}}^{(i)})}_{\text{Lexical}}
$}
\label{eq: retrieve}
\end{equation}
where $\phi(\cdot)$ denotes the semantic text embedding function, and $\oplus$ represents the concatenation operation of multimodal text streams. $\mathcal{K}(\cdot)$ is a lexical scoring function based on keyword hit rate and exact match enhancement, and $\beta$ is a hyperparameter regulating the relative lexical weight. This formulation ensures that the agent possesses dual perception capabilities: the dense vector component captures abstract visual concepts (e.g., \enquote{a warm dinner}), while the sparse retrieval component precisely localizes specific explicit marketing clues (e.g., \enquote{Nike} or \enquote{50\% OFF}).

\begin{table}[t]
    \small
    \centering
    \caption{\textbf{Action Spaces for Intent Reasoning.}
    The framework stratifies the action space into perceptual grounding tools and intent analysis experts to bridge the semantic gap. Specific implementation parameters (e.g., sampling rates) are detailed in \cref{app:Admir Agent Configuration}.}
    \label{tab:action_space}
    
    \begin{tabularx}{\linewidth}{l >{\raggedright\arraybackslash}X}
        \toprule
        \textbf{Action} & \textbf{Function \& Parameters} \\
        \midrule
        \textsc{Global Browse} &
        \textbf{Input:} Database $\mathcal{M}$, Query $Q_{init}$ \newline
        Retrieves pre-computed narrative logs to synthesize a structured global summary $C_{glob}$. \\
        \midrule
        \textsc{Clip Search} &
        \textbf{Input:} Database $\mathcal{M}$, Query $\hat{Q}$ \newline
        Performs hybrid retrieval with Temporal Fusion to merge continuous event segments. \\
        \midrule
        \textsc{Frame Inspect} &
        \textbf{Input:} Query $Q_{vis}$, Range $[t_s, t_e]$, Mode $m$ \newline
        Variable-density visual inspection: $m_{lit}$ (dense) for factual details; $m_{sem}$ (sparse) for stylistic elements. \\
        \midrule
        \textsc{Comm. Expert} &
        \textbf{Input:} Focus $\hat{Q}_{foc}$, Range $[t_s, t_e]$ \newline
        Executes high-density grid inference to resolve long-term dependencies and decode persuasion logic. \\
        \midrule
        \textsc{Finish} &
        \textbf{Input:} Response $A$, Evidence $\mathcal{E}$ \newline
        Triggers $\textsc{VerifyGrounding}$; returns $A^*$ if verified, otherwise initiates a forced backtrack. \\
        \bottomrule
    \end{tabularx}
\end{table}

% ====== 算法：AD-MIR Inference Process ======
\begin{algorithm}[tb]
  \caption{AD-MIR Inference Process}
  \label{alg:admir_inference}
  \begin{algorithmic}
    \STATE {\bfseries Input:} Video Stream $\mathcal{V}$, Query $Q$
    \STATE {\bfseries Output:} Refined answer $A^*$

    \STATE {\bfseries Stage I: Structure-Aware Memory Construction}
    \STATE $\mathcal{M}, \mathcal{S}_{reg} \gets \textsc{BuildDB}(\mathcal{V})$ \COMMENT{1 FPS, Whisper, GPT-4o}

    \STATE {\bfseries Stage II: Iterative Intent Reasoning}
    \STATE $C_{glob}, \text{ASR} \gets \textsc{GlobalBrowse}(\mathcal{M}, Q)$
    \STATE $\mathcal{H}_0 \gets \{Q, C_{glob}, \mathcal{S}_{reg}\}$
    \STATE $t \gets 0$

    \WHILE{$t < T_{\max}$}
      \STATE $z_t, a_t \gets \pi_{\Theta}(\mathcal{H}_{t-1})$ 

      \IF{$a_t = \textsc{Finish}(A, \mathcal{E})$}
        \STATE $\mathcal{W}_{anchors} \gets \textsc{ExtractEntities}(\mathcal{E})$
        \IF{$\textsc{VerifyGrounding}(A, \mathcal{E}, \mathcal{W}_{anchors})$}
          \STATE {\bfseries return} $\textsc{VisualAnchorRefine}(A)$
        \ENDIF
        \STATE $\mathcal{H}_t \gets \mathcal{H}_{t-1} \cup \{\text{"Reject: Weak Evidence"}\}$
        \STATE $t \gets t + 1$
        \STATE {\bfseries continue}
      \ENDIF

      \IF{$a_t \in \mathcal{A}_{percept}$}
        \STATE $o_t \gets \textsc{Execute}(a_t, \mathcal{M})$
      \ELSE
        \IF{$a_t = \textsc{CommExpert}(\hat{Q}_{foc}, [t_s, t_e])$}
          \STATE $L_{vis} \gets \textsc{GridSample}(\mathcal{M}, [t_s, t_e])$
          \STATE $o_t \gets \textsc{Reason}(L_{vis}, \text{ASR}, C_{glob}, \hat{Q}_{foc})$
        \ENDIF
      \ENDIF

      \STATE $\mathcal{H}_t \gets \mathcal{H}_{t-1} \cup \{(z_t, a_t, o_t)\}$
      \STATE $t \gets t + 1$
    \ENDWHILE

    \STATE {\bfseries return} \textsc{Failure}
  \end{algorithmic}
\end{algorithm}

\subsubsection{Context-Anchored Subject Registry} 
In complex advertising scenarios, the frame is often crowded with background characters, which distracts the agent from the main protagonists. To mitigate this noise, we implement a two-step register-and-filter mechanism. 

First, during the offline memory construction, we utilize GPT-4o to extract a static subject registry $\mathcal{S}_{reg} = \{s_j\}_{j=1}^M$. Unlike simple object detection, the model generates a rich semantic profile $d_j$ for each distinct character (e.g., \enquote{A middle-aged doctor wearing a stethoscope and looking anxious}), rather than just a bounding box label.

Second, during inference, we apply a dynamic activation strategy to decide which characters are relevant. User queries often suffer from referential ambiguity (e.g., \enquote{What is \textit{he} holding?}), where the pronoun's target is unclear without context. To resolve this, we construct an enhanced semantic anchor $A_{\text{anchor}}$—a composite representation that fuses the raw query $Q$ with the global narrative summary $C_{glob}$ obtained from the global browse tool ($A_{\text{anchor}} = Q \oplus C_{glob}$). This anchor serves as a contextualized reference point in the vector space, ensuring the search is driven by the full narrative intent rather than ambiguous keywords. We then compute the relevance score $\rho_j$ between this anchor and each character's profile:
\begin{equation}
\resizebox{0.9\linewidth}{!}{%
$\mathcal{S}_{\text{active}} = \text{TopK}\left( \left\{ s_j \in \mathcal{S}_{reg} \mid \rho_j = \cos(\phi(d_j), \phi(A_{\text{anchor}})) \right\}, k \right)$%
}
\end{equation}
By filtering based on semantic similarity $\rho_j$, this mechanism acts as a cognitive spotlight: it activates relevant characters (e.g., the "doctor") while suppressing background noise, ensuring the reasoning chain focuses solely on evidence pertinent to the narrative intent.

\subsection{Structured Reasoning Agent}
At the core of AD-MIR lies a controller designed to emulate the cognitive processes of marketing experts by dynamically orchestrating action primitives.

\subsubsection{Prompt-Guided ReAct Controller.} 
We instantiate the LMM as the central controller $\pi_\Theta$. Distinct from fine-tuned video agents, our system utilizes frozen parameters $\Theta$ (e.g., GPT-4o) governed by structured in-context learning. Instead of relying on implicit priors, the system prompt injects explicit behavioral constraints, mandating that the agent prioritizes retrieved visual evidence over internal parametric knowledge when interpreting abstract concepts. The execution logic follows the "Think-Act-Observe" cycle formalized in \cref{alg:admir_inference}.

At each step $t$, the agent generates a thought $z_t$ and selects an action $a_t$ conditioned on the interaction history $\mathcal{H}_{t-1}$. The controller is equipped with a self-correction mechanism: if the generated answer fails the visual grounding verification check (as detailed in \cref{sec:reliability_mechanisms}), the system triggers a forced trajectory backtrack, appending a \enquote{Weak Evidence} constraint to $\mathcal{H}_t$ to guide the policy toward re-retrieval.

\subsubsection{Search-Centric Toolset Construction.} 
Building upon the structured database $\mathcal{M}$, we categorize the action space $\mathcal{A}$ into two distinct groups: \textit{information retrieval tools} for locating visual evidence and \textit{reasoning modules} for decoding narrative logic. To ensure deterministic interaction between the agent and the environment, we define explicit functional interfaces for each tool, as comprehensively formalized in \cref{tab:action_space}. The implementation details of these tools are elaborated below.

\textbf{Tool: Global Browse.} 
Serving as the strategic initializer for the reasoning trajectory, this tool constructs a holistic narrative framework $C_{glob}$ to ground subsequent actions. In contrast to conventional methods relying on stochastic frame sampling that frequently fractures temporal continuity, this module exploits the pre-computed textual modalities within $\mathcal{M}$. Specifically, it aggregates the complete chronological sequence of visual captions and ASR transcripts into a unified context window. By processing this high-density textual stream, the LMM synthesizes a structured summary that not only classifies the macro-genre (e.g., emotional narrative vs. direct promotion) but also extracts pivotal entities. This global prior effectively prunes the search space by imposing essential semantic constraints, thereby preventing the agent from hallucinating in unrelated contexts.

\textbf{Tool: Clip Search.} 
To secure narrative continuity at the event level, this tool facilitates medium-granularity retrieval, addressing the issue of temporal fragmentation common in standard top-k retrieval. By executing a synthetic query $\hat{Q}$ against the hybrid index, it identifies candidate intervals which are subsequently refined via a Temporal Fusion Algorithm. Specifically, disjointed clips exhibiting minimal temporal gaps (less than 3 seconds) or high semantic affinity ($> 0.8$) are coalesced into unified event blocks. This mechanism effectively reconstructs fragmented retrieval results into continuous event units, ensuring that the agent perceives complete long-duration actions rather than isolated snippets.

\textbf{Tool: Frame Inspect.} 
Functioning as the system's visual microscope, this tool bridges the granularity gap between abstract semantics and raw pixel data by modulating its sampling strategy according to the analysis mode $m$. Specifically, the literal mode ($m_{lit}$) enforces high-frequency dense sampling to rigorously extract observable axioms, such as OCR text and object counts, which are essential for factual verification. In contrast, the semantic mode ($m_{sem}$) adopts a sparse keyframe strategy to interpret stylistic elements like lighting, color temperature, and metaphorical imagery. This dual-path architecture allows the agent to dynamically alternate between forensic precision and aesthetic appreciation based on the specific requirements of the query.

\textbf{Tool: Communication Expert.} 
Designed to bridge the gap between concrete visual signals and abstract persuasive intent, this tool introduces a Spatio-Temporal Grid Projection mechanism. Unlike conventional sequential processing that suffers from context forgetting in long videos, this module transforms the temporal dimension into a unified spatial representation by arranging $N=64$ sampled frames into a high-resolution composite visual grid. This projection enables the LMM to perform parallel visual reasoning over the entire narrative arc simultaneously. Furthermore, to mitigate the hallucinations common in abstract deduction, we implement a Text-Constrained Symbolic Mapping protocol. By fusing the visual grid with aligned ASR and global summaries, the agent is forced to ground its high-level interpretations (e.g., mapping \enquote{soaring eagle} to \enquote{freedom}) in verifiable textual facts, thereby achieving robust persuasion analysis without exceeding token limits.

\subsection{Reasoning Reliability Mechanisms}
\label{sec:reliability_mechanisms}
We embed two control mechanisms within the execution loop to enforce a robust and coherent reasoning trajectory.

\textbf{Macro-to-Micro Evidence Zooming.}
We enforce a hypothesis-driven cascading strategy: 
\begin{equation}
    \mathcal{C}_{\text{glob}} \xrightarrow{\text{Init}} \text{Intent} \xrightarrow{\text{Search}} \mathcal{T}_{rel} \xrightarrow{\text{Verify}} \mathcal{L}_{\text{vis}}. 
\end{equation}
The agent first pre-emptively acquires the narrative contour via \textsc{Global Browse} to initialize the semantic context. Subsequently, within the ReAct loop, it prioritizes the \textsc{Comm. Expert} to formulate high-level persuasion hypotheses (e.g., identifying a specific marketing strategy). These hypotheses guide the \textsc{Clip Search} to narrow down relevant temporal intervals $\mathcal{T}_{rel}$, and finally, the agent locks onto pixel-level evidence through \textsc{Frame Inspect} for rigorous verification. This zooming mechanism prevents disorientation by ensuring that low-level visual search is always directed by high-level semantic intent.

\textbf{Visual Grounding Verification.}
To mitigate hallucinations, we introduce a rigorous answer verification protocol. Upon generating a candidate answer $A_{\text{cand}}$, the system triggers a grounding function $\Psi$:
\begin{equation}
\text{State} = \Psi(A_{\text{cand}}, \mathcal{E}, \mathcal{W}_{\text{anchors}})
\end{equation}
where $\mathcal{E}$ denotes the accumulated evidence chain, and $\mathcal{W}_{\text{anchors}}$ represents the set of visual entities extracted from the retrieving history. The function $\Psi$ checks whether the key claims in $A_{\text{cand}}$ are supported by specific frames in $\mathcal{W}_{\text{anchors}}$. If visual grounding is insufficient (i.e., $\text{State} = \text{Reject}$), the system mandates a forced trajectory backtrack, appending a negative constraint to the history to guide the agent toward alternative retrieval paths.

\begin{table*}[!htbp]
  \centering
\caption{Performance comparison on the AdsQA benchmark. We report both strict and relaxed accuracy (\%) across all five dimensions (VU, ER, TE, PS, AM) and the overall average. AD-MIR, built upon Qwen2.5-VL-7B and o1, is evaluated against commercial LMMs (o1), open-source video LVLMs, and reasoning-oriented video agents.}
\resizebox{\textwidth}{!}{
    \begin{tabular}{lcccccccccccc}
    \toprule
    \multirow{2}[2]{*}{Model} &
    \multicolumn{6}{c}{Strict Accuracy} &
    \multicolumn{6}{c}{Relaxed Accuracy} \\
    \cmidrule(lr){2-7} \cmidrule(lr){8-13}
    & VU & ER & TE & PS & AM & Overall
    & VU & ER & TE & PS & AM & Overall \\
    \midrule

    \rowcolor{row-green}\multicolumn{13}{l}{\gray{\textit{\textbf{Commercial Large Multimodal Model}}}}\\
    \rowcolor{row-green}
    o1\citep{OpenAI2024o1SystemCard}
    &\second{31.8} & 32.0 & 34.9 & 31.0 & 32.0 & 33.1
    &50.9 & 52.8 & \second{54.8} & 50.0 & 51.2 & \second{52.4} \\
    \midrule

    \rowcolor{row-yellow}\multicolumn{13}{l}{\gray{\textit{\textbf{Open-sourced Video-LLMs}}}}\\
    \rowcolor{row-yellow}
    VideoLLaMA2-7B\citep{Cheng2024VideoLLaMA2}
    & 4.56 & 7.75 & 7.28 & 6.06 & 7.38 & 6.48
    & 21.6 & 29.0 & 28.4 & 22.6 & 27.2 & 25.2 \\
    \rowcolor{row-yellow}
    LLaVA-OneVision-7B\citep{Li2024LLaVAOneVision}
    & 11.8 & 11.6 & 16.2 & 13.7 & 15.1 & 14.0
    & 35.8 & 39.5 & 43.2 & 36.8 & 41.7 & 39.1 \\
    \rowcolor{row-yellow}
    LLaVA-Video-7B\citep{zhang2024video}
    & 14.0 & 14.4 & 18.7 & 15.2 & 17.3 & 16.1
    & 37.8 & 41.6 & 45.1 & 38.1 & 43.6 & 41.0 \\
    \rowcolor{row-yellow}
    Qwen2.5-VL-7B\citep{Bai2025Qwen2_5VL}
    & 27.6 & 27.1 & 25.7 & 26.3 & 25.0 & 26.2
    & 49.8 & 51.6 & 48.8 & 48.5 & 47.2 & 49.0 \\
    \midrule

    \rowcolor{row-blue}\multicolumn{13}{l}{\gray{\textit{\textbf{Reasoning Agent/Models}}}}\\
    \rowcolor{row-blue}
    ReAd-R(Qwen2.5-VL-7B)\citep{Long2025AdsQA}
    & 20.4 & 27.9 & 27.2 & 22.1 & 25.5 & 25.0
    & 46.2 & \second{56.2} & 54.6 & 48.2 & \second{52.6} & 51.5 \\
    \rowcolor{row-blue}
    DVD(o1)\citep{Zhang2025}
    & 31.1 & \second{34.0} & \second{39.8} & \second{35.9} & \second{36.6} & \second{36.3}
    & 45.0 & 50.9 & \second{54.8} & 48.5 & 51.5 & 50.5 \\

    \rowcolor{row-pink}
    AD-MIR(Qwen2.5-VL-7B)
    & 31.6 & \second{34.0} & 31.0 & 29.5 & 27.4 & 30.7
    & \second{52.5} & 55.9 & 53.0 & \second{50.2} & 50.1 & 52.1 \\
    \rowcolor{row-pink}
    AD-MIR(o1)
    & \best{32.8} & \best{36.1} & \best{41.6}& \best{37.9} & \best{38.1} & \best{38.1} \gain{1.8}
    & \best{56.0} & \best{59.9} & \best{63.2} & \best{58.8} & \best{60.6} & \best{60.0} \gain{7.6} \\
    \bottomrule
    \end{tabular}%
  }
  \label{tab:main_results}
\end{table*}

\section{Experiments}
In this section, we evaluate AD-MIR on the AdsQA benchmark \citep{Long2025AdsQA}. Our experimental design aims to: (1) benchmark the framework against end-to-end LMMs and general-purpose video agents; (2) isolate the contributions of specific architectural components via ablation studies; and (3) analyze hyperparameter sensitivity to verify system robustness and reproducibility.

\subsection{Experimental Setup}

\textbf{Dataset.}
We evaluate AD-MIR on AdsQA, a benchmark that focuses on deep persuasion logic and marketing intent understanding in advertising videos. AdsQA covers 38 diverse categories and includes questions that range from concrete factual retrieval to abstract rhetorical and audience modelling analysis. This setting precisely matches our goal of testing whether an agent can connect pixel-level evidence to complex high-level persuasion intents.

\textbf{Evaluation Metrics.}
Following the official AdsQA protocol, we employ an LLM-as-a-Judge framework to evaluate semantic alignment. 
The judge receives the video meta-data, query, ground truth, and model prediction, assigning a score $s \in \{0, 0.5, 1.0\}$ based on a standardized rubric (detailed in \cref{app:Evaluation Prompts}). 
We report \textbf{strict accuracy} (fraction of samples with $s=1.0$) and \textbf{relaxed accuracy} ($s \ge 0.5$) across five task-specific dimensions: Visual Concept Understanding (VU), Emotion Recognition (ER), Theme and Core Message Extraction (TE), Persuasion Strategy Mining (PS), and Potential Audience Modeling (AM). 
These metrics allow for a fine-grained diagnosis of different cognitive abilities.
Crucially, to mitigate the well-known length bias of LLM judges and ensure a fair evaluation, we integrate a Refinement Module that strictly constrains our model's output to fewer than 30 tokens, thereby prioritizing information density and precision over verbosity.

\textbf{Baselines.}
We compare AD-MIR against three families of systems, as summarized in \cref{tab:main_results}: (i) a strong commercial LMM (o1) used as an end-to-end zero-shot reference, (ii) recent open-source video LVLMs based on LLaVA-style and Qwen2.5-VL backbones, and (iii) reasoning-oriented video agents, including the AdsQA-specific ReAd-R and the long-video agent DVD. AD-MIR is instantiated on both an open-source backbone (Qwen2.5-VL-7B) and a commercial reasoning model (o1), which we denote as AD-MIR(Qwen2.5-VL-7B) and AD-MIR(o1). Further implementation details for all baselines are provided in \cref{app:Baseline Testing Protocol}.

\begin{figure*}[t]
  \begin{center}
    \centerline{\includegraphics[width=\textwidth]{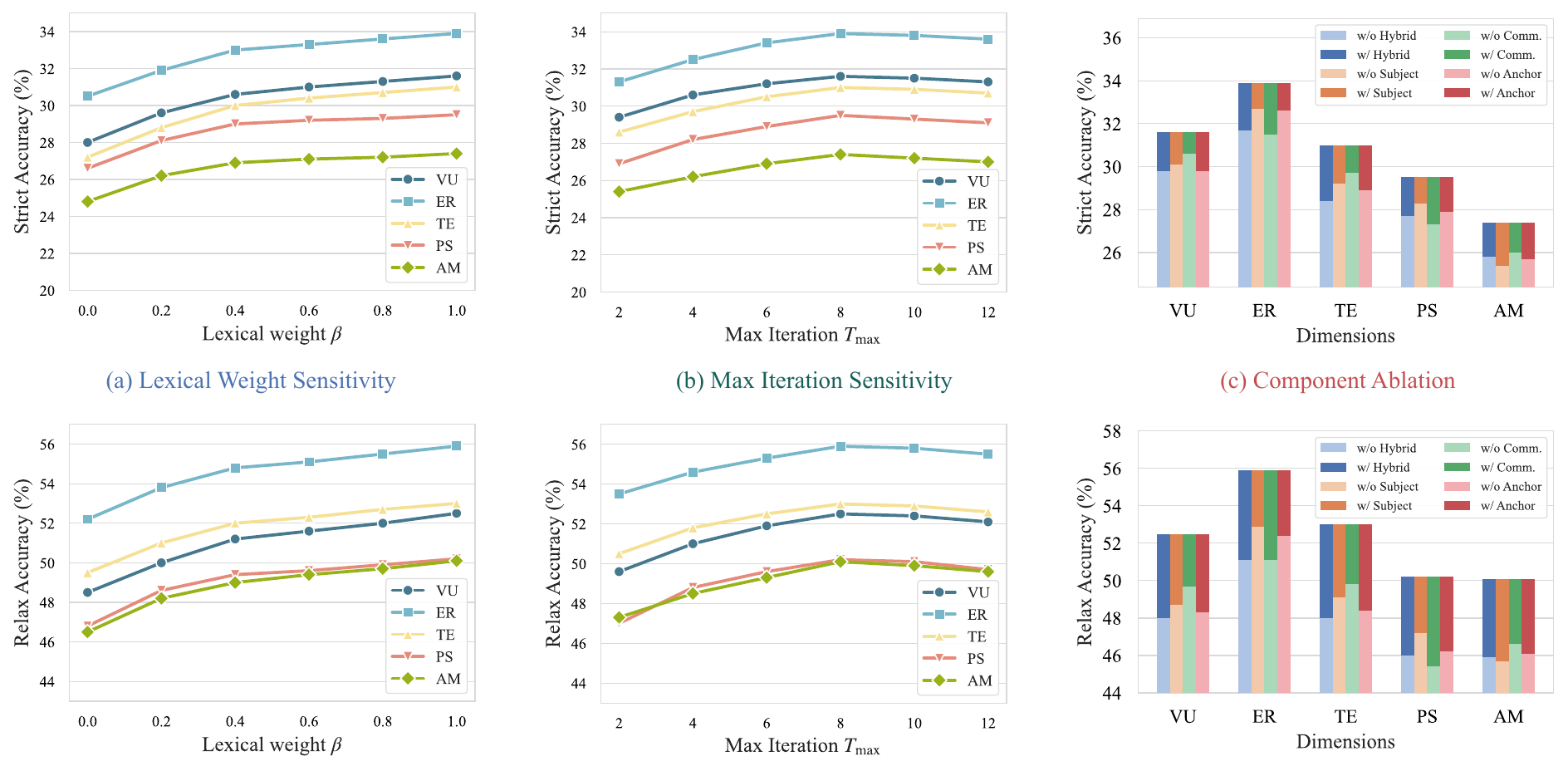}}
    \caption{Ablation and sensitivity analysis of \textbf{AD-MIR} on AdsQA. 
    Subfigures (a) and (b) report strict (top) and relaxed (bottom) accuracy on all five dimensions under different lexical weights $\beta$ and maximum reasoning steps $T_{\text{max}}$, respectively, showing that performance is stable in a broad range and peaks around the default setting. 
    Subfigure (c) presents component-level ablations for Hybrid Indexing, Subject Registry, Communication Expert, and Visual Anchor, where \enquote{w/} and \enquote{w/o} denote whether the corresponding module is enabled; the results highlight the complementary gains of structured indexing, domain expert reasoning, and visual anchor self-correction.}
    \label{fig:ablation}
  \end{center}
\end{figure*}

\subsection{Main Results}
\label{sec:main_results}

We present the comprehensive performance comparison in \cref{tab:main_results}. As evidenced by the data, AD-MIR establishes new state-of-the-art benchmarks across both strict and relaxed accuracy metrics, yielding three critical insights into the efficacy of structured reasoning agents.

\textbf{Unlocking Latent Reasoning Capabilities.}
Our results demonstrate substantial performance gains over base models across both metrics. Equipped with the o1 backbone, AD-MIR achieves a strict accuracy of \textbf{38.1\%} (+5.0\%) and a relaxed accuracy of \textbf{60.0\%} (\textbf{+7.6\%}). Similarly, AD-MIR (Qwen2.5-VL-7B) improves upon its base model by \textbf{+4.5\%} (strict) and \textbf{+3.1\%} (relaxed). These consistent improvements indicate that the bottleneck in complex video understanding lies less in inherent parametric knowledge and more in the orchestration of long-horizon reasoning. By enforcing an explicit \enquote{Think-Act-Observe} loop, AD-MIR effectively unlocks latent deductive capabilities, bridging the gap between raw knowledge and precise decision-making.

\textbf{Bridging the Proprietary-Open Source Gap.}
Crucially, our framework enables smaller open-source models to compete with larger commercial systems. AD-MIR (Qwen2.5-VL-7B) attains a relaxed accuracy of \textbf{52.1\%}, comparable to the commercial o1 baseline (52.4\%). This suggests that a structured agentic architecture augmented with domain-specific tools (e.g., the \textit{Communication Expert}) can mitigate parameter disparity, providing a resource-efficient alternative for high-level video reasoning.

\textbf{Superiority Over Reasoning-Enhanced Models.}
AD-MIR also surpasses reasoning-centric baselines like ReAd-R. While ReAd-R shows strong latent reasoning (51.5\% relaxed), its strict accuracy collapses to 25.0\%, highlighting the difficulty of aligning reasoning with rigid output constraints through standard optimization. Conversely, AD-MIR (Qwen2.5-VL-7B) achieves a favorable balance (+5.7\% strict, +0.6\% relaxed) without parameter updates. This validates our training-free paradigm: instead of relying on potentially unstable fine-tuning, AD-MIR leverages structured tool usage to guide the frozen backbone, proving that architectural scaffolding offers a more robust mechanism for enforcing verifiable decision-making.

\subsection{Ablation Study and Sensitivity Analysis}
\label{sec:ablation}

We conduct ablation and sensitivity analyses using Qwen2.5-VL-7B to isolate component contributions and evaluate system robustness, as summarized in \cref{fig:ablation}. The results validate our architectural choices.

\textbf{Sensitivity to Hyperparameters.}
\cref{fig:ablation}(a) examines the impact of the lexical weight $\beta$ in our hybrid indexing mechanism. Performance improves significantly as $\beta$ increases and stabilizes across a broad range of higher values. This trend underscores the critical role of precise lexical matching for AdsQA, where intent often hinges on fine-grained OCR clues (e.g., disclaimers, prices) that semantic-only retrieval might overlook.
\cref{fig:ablation}(b) investigates the maximum reasoning steps $T_{\max}$. Accuracy improves with increased reasoning budgets up to $T_{\max}=8$, confirming the necessity of multi-hop evidence collection for complex narratives. Beyond this point, performance saturates and even slightly declines, suggesting that excessive reasoning steps yield diminishing returns and may introduce noise.

\textbf{Component-wise Ablation.}
\cref{fig:ablation}(c) validates the efficacy of our architectural modules by removing them one at a time:
(1) \textbf{w/o Hybrid Indexing}: Removing lexical signals leads to a notable performance drop, reaffirming that semantic retrieval alone is insufficient for capturing exact visual text details.
(2) \textbf{w/o Subject Registry}: The exclusion of subject tracking degrades performance, particularly in questions involving social interactions, due to increased interference from irrelevant background characters.
(3) \textbf{w/o Communication Expert}: We observe the sharpest decline in persuasion-centric dimensions (e.g., PS) when this module is disabled, supporting our hypothesis that explicit marketing priors are essential for decoding advertising strategies.
(4) \textbf{w/o Visual Anchor}: The absence of the visual anchor consistently hurts strict accuracy, indicating that the verification mechanism is crucial for suppressing plausible but hallucinated reasoning.

\section{Conclusion}
\label{sec:conclusion}
In this paper, we presented AD-MIR, a multimodal framework designed to bridge the cognitive gap between pixel-level perception and high-level persuasive intent. By synergizing structure-aware memory construction with iterative structured reasoning, AD-MIR effectively decodes complex marketing strategies within non-linear narratives. Extensive evaluations on the AdsQA benchmark demonstrate that our approach achieves state-of-the-art performance, significantly outperforming both end-to-end LMMs and general video agents. These findings highlight the necessity of explicit reasoning for advertising understanding and pave the way for next-generation agents capable of strategic analysis.

\section*{Impact Statement}

This paper presents work aimed at advancing multimodal video understanding, specifically in the domain of advertising and persuasion analysis. A primary societal benefit of our framework, AD-MIR, is its potential to assist in content moderation and regulatory compliance by automatically decoding implicit marketing intents and detecting misleading or non-compliant persuasive tactics. This capability contributes to greater transparency in digital media.

% In the unusual situation where you want a paper to appear in the
% references without citing it in the main text, use \nocite
\nocite{langley00}

\bibliography{example_paper}
\bibliographystyle{icml2026}

\newpage
\appendix
\crefalias{section}{appendix}
\crefalias{subsection}{appendix}
\onecolumn
\newcommand{\K}[2]{%
    \vspace{0.3em}\noindent%
    \tikz[baseline=(X.base)]{
        \node[
            fill=#1!12,      
            text=#1!80!black, 
            rounded corners=3pt,
            inner sep=3pt,
            anchor=text
        ] (X) {\sffamily\bfseries #2};
    }%
    \hspace{0.3em}%
}

\newtcolorbox{promptbox}[1]{
    enhanced, breakable,
    title={#1},
    fonttitle=\sffamily\bfseries\large,
    coltitle=white,
    attach boxed title to top left={xshift=6mm, yshift=-3.5mm},
    boxed title style={
        colback=primaryColor,
        arc=4pt,   
        boxrule=0pt,
        drop fuzzy shadow=black!30
    },
    colframe=primaryColor,
    colback=bgColor,
    coltext=textColor,
    boxrule=0.5pt,
    arc=3mm,
    left=4mm, right=4mm, top=6mm, bottom=4mm,
    fontupper=\sffamily\small,
    drop fuzzy shadow,
    overlay={
        \draw[primaryColor, line width=2pt] 
        ([yshift=-3mm]frame.north west) -- ([yshift=3mm]frame.south west);
    }
}

\newtcblisting{codebox}{
    enhanced,
    colback=codeBg,
    colframe=gray!20,
    boxrule=0.5pt,
    arc=2mm,
    left=2mm, right=2mm, top=2mm, bottom=2mm,
    listing only, 
    listing options={
        basicstyle=\ttfamily\footnotesize\color{textColor}, % 字体设置
        breaklines=true,      
        columns=fullflexible,   
        keepspaces=true,       
        language=Java,          
        keywordstyle=\color{primaryColor},  
        stringstyle=\color{accentColor!80!black},    
        showstringspaces=false  
    },
    borderline west={2pt}{0pt}{gray!40}
}
\newtcblisting{contentbox}{
    enhanced,
    colback=codeBg,
    colframe=gray!20,
    boxrule=0.5pt,
    arc=2mm,
    left=2mm, right=2mm, top=2mm, bottom=2mm,
    listing only,
    listing options={
        basicstyle=\ttfamily\scriptsize\color{textColor},
        breaklines=true,
        columns=fullflexible,
        keepspaces=true,
        showstringspaces=false,
        frame=none
    },
    borderline west={2pt}{0pt}{gray!40}
}

\section{Related Works}

\textbf{Advertising Understanding}
Computational advertisement understanding has long recognized that ads are deliberately engineered persuasive artifacts rather than objective records, making the key challenge to infer persuasive intent beyond literal content. \citet{Hussain2017} formalize this view by connecting observable cues (objects, scenes, edits) to symbolic references and persuasive strategies in large-scale image and video ad datasets. Building on this semiotic framing, our work focuses on modern long-form, information-dense advertising videos and emphasizes intent reasoning that leverages domain knowledge (marketing psychology, rhetoric) together with stricter evidence grounding to avoid plausible-but-unsupported narratives.

\textbf{Long-Form Video Understanding with Multimodal LLMs and Benchmarks} 
While the field of visual generation has witnessed rapid advancements in controllable autoregressive learning \citep{xu2025scalar, jin2025semantic}, precise scene text editing \citep{lan2025flux}, and high-fidelity video synthesis applications \citep{wu2025hunyuanvideo, zeng2025eevee}, the domain of \textit{understanding} long-form narratives faces distinct challenges.
Despite rapid progress in multimodal LLMs for video description and QA, long-form understanding remains bottlenecked by limited context, temporal fragmentation, and imperfect multimodal integration. Representative advances include unified frameworks like Video-LLaVA \citep{Lin2024} and memory-augmented models such as MA-LMM \citep{He2024MALMM} and PAM~\cite{lin2025perceive}. However, recent benchmarks, including Video-MME, LVBench, ALLVB, MMBench-Video and  MME-CoF \citep{Fu2025,Wang2025LVBench,Tan2025ALLVB,Fang2024MMBenchVideo,guo2025video}, collectively reveal that current systems still degrade markedly on multi-hour videos and require explicit long-term memory. Motivated by these findings, our approach treats advertising videos as high-density long-form narratives, incorporating structured memory, temporal indexing, and iterative evidence collection tailored to intent-centric reasoning.

\textbf{Tool-Augmented Agents and Hallucination-Aware Multimodal Reasoning}
A promising solution to long-context constraints is tool-augmented, agentic reasoning that iteratively gathers evidence and revises hypotheses. On the language side, \citet{Yao2023,Schick2023Toolformer} show that LLMs can interleave reasoning with external tool calls and even self-supervise tool-use policies, while in the video domain \citet{Zhang2025,Fan2024VideoAgent} develop agentic frameworks such as DVD and VideoAgent for multi-granular retrieval and memory-augmented long-horizon understanding, and hallucination analyses including HallusionBench and object-centric evaluations like POPE and Logical Closed Loop by \citet{Guan2024,Li2023POPE,Wu2024LogicalLoop} demonstrate that LVLMs frequently produce confident yet weakly supported explanations. Building on this line of work, our method instantiates an advertising-specific tool-augmented agent with domain-adaptive inductive biases and hallucination-aware self-correction, explicitly enforcing evidence-first reasoning when mapping pixel-level anchors to persuasion-level intents.

\section{Implementation Details}
\label{app:implementation_details}

In this section, we provide a detailed overview of the experimental setup, including the testing protocols for baseline models and the specific hyperparameter configurations for our proposed \textsc{Admir} agent.

\subsection{Baseline Testing Protocol}
\label{app:Baseline Testing Protocol}
Following the benchmarking standards established in AdsQA, we evaluate baseline Multimodal Large Language Models (MLLMs) using a standardized zero-shot setting to ensure fair comparison.

\textbf{Frame Sampling Strategy.} For end-to-end MLLM baselines, we adopt a uniform sampling strategy. Given the high temporal redundancy in advertisement videos, we sample $N=16$ frames uniformly distributed across the video duration. For models that support high-resolution inputs or dynamic resolution (e.g., o1), frames are resized while maintaining the aspect ratio, with the longest side not exceeding 1024 pixels.

\textbf{Prompting Strategy.} We utilize the standard prompts provided by the AdsQA benchmark. The prompt structure consists of a system instruction defining the task (Video QA), followed by the visual inputs (interleaved frames) and the specific question. For open-ended generation, we employ a greedy decoding strategy to ensure reproducibility.

\subsection{Admir Agent Configuration}
\label{app:Admir Agent Configuration}
The implementation details for the data preprocessing and agent hyperparameters are strictly aligned with the provided logic.

\textbf{Video Database Construction.}
We preprocess the AdsQA video corpus to build a structured retrieval database:
\begin{itemize}
    \item \textbf{Frame Extraction}: Videos are decoded at a frame rate of 1 FPS (`VIDEO\_FPS=1`).
    \item \textbf{Captioning}: We generate dense captions for video clips with a duration of 5 seconds (`CLIP\_SECS=5`). The captioning model is \texttt{gpt-4o}.
    \item \textbf{Subject Registry}: A dynamic subject registry is maintained and merged across clips to track character identities and key objects.
    \item \textbf{Embeddings}: We utilize \texttt{BAAI/bge-m3} (dimension 1024) for local vector storage.
\end{itemize}

\textbf{Agent Hyperparameters.}
The \textsc{Admir} agent is orchestrated by \texttt{gpt-4o}, while the specialized \textsc{Communication Expert} tool utilizes the reasoning-intensive \texttt{o1} model.
\begin{itemize}
    \item \textbf{Maximum Iterations}: The ReAct loop is capped at $T_{max}=8$ steps to prevent infinite loops, though most queries are resolved within 4--6 steps.
    \item \textbf{Global Browse}: The top $K=40$ semantic captions (`GLOBAL\_BROWSE\_TOPK`) are retrieved during the pre-emptive global browse phase to construct the initial narrative context.
    \item \textbf{Clip Search}: A hybrid search mechanism (Vector + Keyword) retrieves the top $K=5$ to $8$ relevant clips for fine-grained inspection.
    \item \textbf{Expert Analysis}: The \textsc{Communication Expert} processes up to 64 frames stitched into $2\times2$ grid images to analyze visual rhetoric and persuasion strategies.
    \item \textbf{Answer Refinement}: The final output is refined by \texttt{gpt-4o-mini} to strictly adhere to the token limit (under 30 words) while preserving critical entities and attributes.
\end{itemize}

\textbf{Retry Mechanism.} To handle API instability and potential hallucinations in tool calls, we implement an exponential backoff retry mechanism with a maximum of 8 retries for critical API failures.

\subsection{Algorithm Robustness Details}
\label{sec:robustness}

To prevent the ReAct agent from falling into infinite loops or overwriting high-confidence expert insights with low-level hallucinations, we implement three specific algorithmic safeguards within the inference controller.

\textbf{1. Temporal Stagnation Detection \& Redirection.} 
In long-horizon reasoning, agents often become fixated on a specific time segment, repeatedly searching the same interval despite finding no new evidence. To mitigate this, we implement a \textit{Temporal Stagnation Check}.
Let $\mathcal{H}_{time} = \{(t_{start}^{(i)}, t_{end}^{(i)})\}$ be the history of time ranges queried by the agent. For a new query range $[t_s, t_e]$ with duration $d = t_e - t_s$, we calculate the overlap ratio with history:
\begin{equation}
    Overlap(i) = \frac{\text{intersection}([t_s, t_e], [t_{start}^{(i)}, t_{end}^{(i)}])}{d}
\end{equation}
If $Overlap(i) > 0.6$ for more than $N=2$ distinct historical queries, the system triggers a \textbf{Forced Redirection Intervention}. The agent's query is overridden, and the focus is mechanically shifted to the video boundaries (First 15s or Last 15s) to break the cognitive deadlock.

\textbf{2. Anti-Verification Protocol (The "Golden Source" Rule).}
A common failure mode in hierarchical agents is "over-verification," where the agent uses a low-capacity tool (e.g., \texttt{frame\_inspect}) to verify a high-capacity tool (e.g., \texttt{communication\_expert}), leading to false negatives due to sparse sampling. 
We enforce an \textit{Anti-Verification Protocol} via the system prompt:
\begin{quote}
    \textit{"IF the Communication Expert returns a specific Proper Name, Metaphor, or Strategy, YOU MUST ACCEPT IT AS FACT. DO NOT use frame\_inspect\_tool to 'double check' what the Expert Tool has already found."}
\end{quote}
This ensures that high-level semantic findings (e.g., "The ad uses a fear appeal strategy") are not discarded simply because a literal frame inspection failed to OCR the word "fear".

\textbf{3. Visual Anchor Repair.}
For questions explicitly demanding visual grounding (e.g., "What specific object..."), the agent's draft answer is passed through a lightweight verification module (using GPT-4o-mini). If the answer lacks concrete visual nouns (anchors) found in the retrieval evidence, a repair loop is triggered to inject specific observable details (e.g., changing "a vehicle" to "a red Ferrari") before final output.

\subsection{Expert Model Input Strategy}
\label{sec:expert_input}

To bridge the gap between token limits and the need for high temporal resolution in advertising analysis, we employ a \textbf{Spatio-Temporal Grid Projection} strategy for the \texttt{communication\_expert\_tool}.

\textbf{High-Density Sampling.} Unlike standard approaches that sample 8-16 frames, our expert module samples $N=64$ frames uniformly distributed across the video duration (or the queried segment). This high density is crucial for capturing fleeting subliminal cues typical in commercials (e.g., split-second micro-expressions or flash cuts).

\textbf{2x2 Grid Stitching.} Directly feeding 64 images exceeds the file count limits of most VLM APIs. We therefore stitch consecutive frames into $2 \times 2$ grid images.
\begin{itemize}
    \item \textbf{Batching}: The 64 frames are divided into 16 batches of 4 frames each.
    \item \textbf{Composition}: Each batch is stitched into a single high-resolution image ($H_{grid} = 2 \times H_{frame}, W_{grid} = 2 \times W_{frame}$).
    \item \textbf{Resolution}: The resulting grids are encoded with \texttt{detail: "high"} to preserve OCR-readability in sub-frames.
\end{itemize}

\textbf{Temporal Serialization.} The VLM receives a sequence of 16 grid images. To ensure correct temporal reasoning, the system prompt explicitly defines the reading order:
\begin{equation}
    \text{Order: } \text{Top-Left} \xrightarrow{} \text{Top-Right} \xrightarrow{} \text{Bottom-Left} \xrightarrow{} \text{Bottom-Right}
\end{equation}
This projection allows the model to process 64 frames of visual information while only consuming the request overhead of 16 images, effectively quadrupling the temporal context window.

\section{Core Prompts of AD-MIR}
In this section, we present the comprehensive system prompts used in the AD-MIR framework. These prompts are meticulously designed to guide the neuro-symbolic agent through the distinct stages of perception, reasoning, and answer refinement.

\subsection{Perception \& Context Construction Prompts}
The following prompts drive the offline perception layer, transforming raw video signals into structured semantic indices through narrative reconstruction and dynamic inspection.

\begin{promptbox}{Global Browse: Text-Based Forensic Reconstruction}
    \K{primaryColor}{Role} Media Forensics Expert
    
    \K{primaryColor}{Goal} Reconstruct the narrative and ground truth from multimodal logs (Visual Captions + Audio) without direct pixel access.

    \K{accentColor}{Key Mechanisms}
    \begin{itemize}
        \item \textbf{A/V Cross-Referencing}: Combining visual descriptions with audio transcripts to infer specific entities (e.g., "Product shown" + "Audio mentions Pepsi" $\to$ "Pepsi").
        \item \textbf{Text Hunting}: Specifically scanning logs for \texttt{[POTENTIAL\_TEXT]} tags to extract explicit on-screen text.
    \end{itemize}

    \K{black}{System Prompt}
    \begin{contentbox}
You are a Media Forensics Expert specialized in solving AdsQA cases.
**YOUR CONSTRAINT**: You cannot see the video. You only have access to:
1. **Visual Logs**: Text descriptions of what happens.
2. **Audio Transcript**: What was said.

**STRATEGY FOR TEXT-ONLY ANALYSIS**:
1. **Cross-Referencing**: If Visual Log says "[SCENE]: A product is shown" and Audio says "Try the new Pepsi", you must infer the product is "Pepsi".
2. **Text Hunting**: Look specifically for the `[POTENTIAL_TEXT]` tags in the logs.
3. **Implicit Clues**: Identify brands/objects from specific descriptions (e.g., red car + horse logo $\to$ Ferrari).

**OUTPUT FORMAT (JSON)**:
{
  "narrative_reconstruction": "Story flow (Hook $\to$ Middle $\to$ End).",
  "inferred_objects": ["List of objects implied by combining visual + audio"],
  "explicit_text_found": ["Text explicitly quoted in visual logs"],
  "audio_visual_mismatch": "Contradictions between seen and heard?",
  "final_answer": "Direct answer to the USER QUERY."
}
    \end{contentbox}
\end{promptbox}

\begin{promptbox}{Frame Inspect: Dynamic Mode Switching}
    \K{primaryColor}{Function} `frame\_inspect\_tool`
    
    \K{primaryColor}{Strategy} The tool dynamically selects an analysis mode based on the query type (Factual vs. Abstract).

    \K{black}{Mode A: Literal Inspection (OCR \& Factual)}
    \begin{contentbox}
[MODE: LITERAL INSPECTION & OCR]
1. **VISUAL LOG**: Create a chronological log of events.
2. **TEXT**: Transcribe any visible text/logos verbatim.
3. **DETAILS**: List objects and specific physical interactions.
    \end{contentbox}

    \K{black}{Mode B: Semantic Analysis (Narrative \& Symbolism)}
    \begin{contentbox}
[MODE: SEMANTIC & NARRATIVE]
1. **STORY**: Describe the setup -> action -> outcome.
2. **TWIST**: Note if the environment changes artificially (e.g., reveal of a set).
3. **MEANING**: Identify key symbols and subject relationships.
    \end{contentbox}
\end{promptbox}

\begin{promptbox}{Clip Search: Semantic Query Rewrite}
    \K{primaryColor}{Function} `clip\_search\_tool`
    
    \K{primaryColor}{Goal} Generalize specific user queries into broader visual concepts for better retrieval recall.

    \K{black}{Rewrite Prompt}
    \begin{contentbox}
Rewrite '{original_query}' into 3 simple, comma-separated keywords/phrases for video retrieval.
Rules: Keep it generic. Do NOT invent specific visual details (e.g. do NOT change 'luxury' to 'gold').
Input: 'sad man' -> Output: crying person, unhappy face, depressed mood
    \end{contentbox}
\end{promptbox}

\subsection{Structured Agent Reasoning Prompts}
These prompts control the online inference engine, managing the interaction between the ReAct controller and domain-specific experts.

\begin{promptbox}{Communication Expert: Visual Semiotics Analysis}    
    \K{primaryColor}{Input} 2x2 Grid Images (High-Density Temporal Sampling)
    
    \K{accentColor}{Core Analysis Protocols}
    \begin{itemize}
        \item \textbf{2x2 Grid Reading}: Explicit instruction on reading order (TL $\to$ TR $\to$ BL $\to$ BR).
        \item \textbf{Grounding}: Strict prohibition against hallucinating objects not present in the grid.
    \end{itemize}

    \K{black}{System Prompt}
    \begin{contentbox}
You are an Elite Advertising Forensics Expert & Visual Semiotics Analyst.
**YOUR MISSION**: Decode the provided video content to uncover the narrative structure, character relationships, and persuasive strategy.

**INPUT FORMAT (CRITICAL)**:
- The visual input consists of **2x2 Grid Images**.
- Each image contains **4 chronological video frames**.
- **Reading Order**: Analyze each grid from **Top-Left -> Top-Right -> Bottom-Left -> Bottom-Right**.

**CORE ANALYSIS PROTOCOLS:**
1.  **OCR & BRAND TRUTH (HIGHEST PRIORITY)**: Any text on screen is fact. Identify Logos.
2.  **UNIVERSAL CHARACTER DYNAMICS**: Analyze Transactional, Conflict, or Affectionate interactions.
3.  **NARRATIVE ARC**: Hook -> Problem -> Product -> CTA.
4.  **GROUNDING**: Do not invent objects not present in the grids.
    \end{contentbox}
\end{promptbox}

\begin{promptbox}{ReAct Controller: System Instructions}
    \K{primaryColor}{Role} Advanced Video Analysis Agent
    
    \K{primaryColor}{Philosophy} "Trust Your Expert \& Be Decisive"

    \K{black}{System Prompt}
    \begin{contentbox}
You are an advanced Video Analysis Agent. Your goal is to answer the user's question precisely using the provided tools.

**CORE PHILOSOPHY: TRUST YOUR EXPERT & BE DECISIVE**
1. **THE "GOLDEN SOURCE" RULE**: The `communication_expert_tool` is your PRIMARY SOURCE OF TRUTH. If it returns specific Proper Names or Metaphors, YOU MUST ACCEPT IT AS FACT.
2. **ANTI-VERIFICATION PROTOCOL**: DO NOT use `frame_inspect_tool` to "double check" the Expert Tool. Only use it as a fallback.
3. **SINGLE-SHOT VICTORY**: If the Expert Tool answers fully, FINISH IMMEDIATELY.

**OUTPUT PROCESS**:
1. **THOUGHT**: First, write a brief thought analysis in plain text explaining your reasoning.
2. **ACTION**: Then, call the appropriate tool using the native function calling capability.
    \end{contentbox}
\end{promptbox}

\begin{promptbox}{Answer Refinement \& Repair}
    \K{primaryColor}{Stage} Post-Processing
    
    \K{primaryColor}{Task} Compress output to $<$ 30 tokens and repair visual grounding errors.

    \K{black}{Refinement Prompt}
    \begin{contentbox}
You are compressing an answer for a visual QA benchmark.
Rewrite the answer to be <= 25 words, but DO NOT lose any core information.

[HARD CONSTRAINTS]
- Preserve ALL essential facts: names/entities, key attributes (color, number, time), and negation.
- DO NOT add new facts or hallucinations.
- If original is entity identification, output a compact noun phrase.

[EDITING RULES]
1) Remove meta phrases: "the video shows", "the answer is".
2) Keep specificity: numbers, brand names, proper nouns.
    \end{contentbox}

    \K{black}{Visual Grounding Repair Prompt}
    \begin{contentbox}
Rewrite the answer so it names ONE specific, observable scene/object/action from the evidence.
Hard constraints:
- Mention a concrete visual element (what is shown), not abstract traits.
- Use at least 1 anchor word if possible.
- Under 30 tokens.
    \end{contentbox}
\end{promptbox}

\subsection{Data Preprocessing Prompts}
The following prompts drive the offline perception layer, transforming raw video signals into structured semantic indices through narrative reconstruction, entity tracking, and fine-grained text extraction.

\begin{promptbox}{Visual Captioning \& Subject Registry Extraction}
    \K{primaryColor}{Input} Consecutive Video Frames (Clip Level)
    
    \K{accentColor}{Core Analysis Protocols}
    \begin{itemize}
        \item \textbf{Structured Output}: Strict JSON format enforcement for database indexing.
        \item \textbf{Subject Profiling}: Extract detailed appearance and identity for the context-anchored registry.
    \end{itemize}

    \K{black}{System Prompt}
    \begin{contentbox}
Here are consecutive frames from a video clip. Please visually analyze the video clip and output JSON in the template below.

Output template:
\{
  "clip\_start\_time": CLIP\_START\_TIME,
  "clip\_end\_time": CLIP\_END\_TIME,
  "subject\_registry": \{
    "<subject\_i>": \{
      "name": "<fill with short identity if name is unknown, e.g. 'man in red'>",
      "appearance": "<list of visual appearance descriptions>",
      "identity": "<list of inferred identity descriptions>",
      "first\_seen": "<timestamp>"
    \},
    ...
  \},
  "clip\_description": "<smooth and detailed visual narration of the video clip>"
\}
    \end{contentbox}
\end{promptbox}

\begin{promptbox}{Subject Registry Merging}
    \K{primaryColor}{Input} List of Partial Subject Registries (from multiple clips)
    
    \K{accentColor}{Core Analysis Protocols}
    \begin{itemize}
        \item \textbf{De-duplication}: Merge subjects referring to the same visual entity across time.
        \item \textbf{Union}: Combine attribute fields while preserving the earliest timestamp.
    \end{itemize}

    \K{black}{System Prompt}
    \begin{contentbox}
You are given several partial `new\_subject\_registry` JSON objects extracted from different clips of the *same* video.

Task:
1. Merge these partial registries into one coherent `subject\_registry`.
2. Preserve all unique subjects.
3. If two subjects visually refer to the same person/object, merge them (keep earliest `first\_seen` time and union all fields).

Input (list of JSON objects):
REGISTRIES\_PLACEHOLDER

Return *only* the merged `subject\_registry` JSON object.
    \end{contentbox}
\end{promptbox}

\begin{promptbox}{Fine-Grained OCR Extraction}
    \K{primaryColor}{Input} Single Video Frame (High Resolution)
    
    \K{accentColor}{Core Analysis Protocols}
    \begin{itemize}
        \item \textbf{Exhaustive Extraction}: Capture all visible text including logos, brand names, and background signs.
        \item \textbf{Format}: One item per line, strictly no visual descriptions.
    \end{itemize}

    \K{black}{System Prompt}
    \begin{contentbox}
Extract ALL visible text from this image. 
Include: titles, labels, captions, signs, logos, brand names, slogans, any written content.
Return ONLY the extracted text, one item per line.
If no text is visible, return "NO\_TEXT".
Do not describe the image, only extract text.
    \end{contentbox}
\end{promptbox}

\subsection{Evaluation Prompts}
\label{app:Evaluation Prompts}
The prompt utilized by the LLM-as-a-Judge mechanism to benchmark performance against ground truth annotations.

\begin{promptbox}{AdsQA Evaluation Judge}
    \K{primaryColor}{Role} Advertising Expert Judge
    
    \K{primaryColor}{Goal} Evaluate semantic alignment between prediction and golden answer.

    \K{black}{Judge Prompt}
    \begin{contentbox}
You are an advertising expert specializing in evaluating whether a respondent's answer after watching a video matches the golden answer.

### The meta-information (Ground Truth):
{meta_info}

### Question: 
{question}

### Golden Answer: 
{golden_answer}

### Rule:
1. If response contains ALL key info OR expresses same meaning -> Output 1.
2. If response does NOT contain key info -> Output 0.
3. If response is contradictory/unreasonable given meta-info -> Output 0.
4. If response contains MOST key info and NO contradictions -> Output 0.5.

### Response to be judged: 
{response}

### Instructions:
Follow the format below and do not give any extra outputs:
Answer: 0 / 0.5 / 1
    \end{contentbox}
\end{promptbox}

\end{document}